%% file: main.tex
\documentclass{article} 
\usepackage{iclr2023_conference,times}

\input{math_commands.tex}

\usepackage[utf8]{inputenc}
\usepackage{url}
\usepackage{graphicx}
\usepackage{listings}
\usepackage{hyperref}

\title{Synthesis of Mathematical Programs from Natural Language Specifications}
\author {Ganesh Prasath Ramani \& Shirish Karande\\
Tata Consultancy Services Research, India\\
\{ganesh.prasathr,shirish.karande\}@tcs.com
}
\date{January 2023}
\begin{document}
\maketitle
\begin{abstract}
\input{abstract}  
\end{abstract}
\section{Introduction} \label{Introduction}
\input{introduction.tex}
\section{Related Work} \label{Related Work}
\input{relatedwork.tex}
\section{Preliminaries} \label{prelims}
\subsection{NL4OPT}
\input{data}

\section{Methodologies for Fine-tuned models} \label{Methodologies}
\input{setup}

\subsection{Post Inference processing}
\input{postprocess}

\subsection{Zero shot synthesis using Codex and ChatGPT}
\input{zeroshot}
\section{Results} \label{Results}
\input{results}
\section{Conclusion} \label{conclusion}
\input{conclusion}
\nocite{*}
\bibliography{ref}
\bibliographystyle{iclr2023_conference}
\newpage
\section{Appendix}
\input{Appendix.tex}
\end{document}

%% file: math_commands.tex
\usepackage{amsmath,amsfonts,bm}

\def\eqref#1{equation~\ref{#1}}

\def\1{\bm{1}}

\DeclareMathAlphabet{\mathsfit}{\encodingdefault}{\sfdefault}{m}{sl}
\SetMathAlphabet{\mathsfit}{bold}{\encodingdefault}{\sfdefault}{bx}{n}



%% file: abstract.tex
Several decision problems that are encountered in various business domains can be modeled as mathematical programs, i.e. optimization problems. The process of conducting such modeling often requires the involvement of experts trained in operations research and advanced algorithms. Surprisingly, despite the significant advances in the methods for program and code synthesis, AutoML, learning to optimize etc., there has been little or no attention paid to automating the task of synthesizing mathematical programs. We imagine a scenario where the specifications for modeling, i.e. the objective and constraints are expressed in an unstructured form in natural language (NL) and the mathematical program has to be synthesized from such an NL specification. In this work we evaluate the efficacy of employing CodeT5 with data augmentation and post-processing of beams. We utilize GPT-3 with back translation for generation of synthetic examples. Further we apply rules of linear programming to score beams and correct beams based on common error patterns. We observe that with these enhancements CodeT5 base gives an execution accuracy of 0.73 which is significantly better than zero-shot execution accuracy of 0.41 by ChatGPT and 0.36 by Codex.

%% file: introduction.tex
Management of budget portfolio across various advertisement channels, procurement of raw materials and machinery in textile production, allocation of resources to a software or engineering project, distribution of vaccines and vaccination centers, etc., all these decision problems and many more across a variety of business and application domains can be modeled as optimization problems. Appropriate modeling can provide an immediate impact on profitability, safety, sustainability etc. In essence achieving maximum benefit from limited resources is the key for every successful business. However, typically a team of experts is required to identify the right decision variables, constraints and objectives in order to formulate the problem and solve it using a optimization solver.
Therefore, only critical operations are given priority and optimized, leaving a large number of decisions unmodeled and therefore not optimized.  Consequently automation in formulation of optimization problems can democratize the process and allow inclusion of broader set of decision variables and decision makers.

Recent years have seen significant advances in the use of neural networks for synthesis of both procedural e.g \citep{PLBART}, \citep{CodeT5} and \citep{karelio} as well as declarative e.g. \citep{FOLIO} and \citep{seq2sql} programs. Optimization problems are similar to declarative programs but expressed with constraints and objectives. Just as done with declarative programs the execution is not to be determined by the developer; typically off the shelf solvers will be utilized and the developer needs to only specify the type of the solver to be used. Its possible to express declarative programs in high level languages like python, for example \citep{pylog}, \citep{KIMMIG_2011} are libraries used for logic programs,  meanwhile \citep{Mitchell2011PuLPA}, \citep{CVXPY} are examples of libraries used for optimization problems.

Thus one who looks to autoformulate optimization problems may also desire automatic synthesis of code in a high level language using a library like PuLP or CVXPY. Nevertheless, despite these similarities there is a surprising dearth of work on automatic synthesis mathematical programs and the relevant code. The recent competition \citep{NL4OPT} proposed at Neurips 22 is a notable exception. 

In this paper, we envision a scenario where a business problem is communicated through natural language, and an autoformulation synthesizer must recognize the constraint and objectives. On account of a dearth of publicly available benchmark datasets we focus on the NL4OPT dataset which has defined tasks to help convert word problems to canonical form expression of linear programs.  We show that by using proposed post processing and training methods we were able to achieve better canonical accuracy of 89.63 with a smaller model (CodeT5-base) compared to the larger (BART-large) model used by current SOTA  88.20 with lesser training epochs (30 vs 400).

The remainder of the paper is organized as follows: Section \ref{Related Work} summarizes the related work in literature. Section \ref{prelims} covers the preliminaries, providing details about the NL4OPT dataset and the metrics of performance employed in this work. Section \ref{Methodologies} describes the details of data augmentation and post processing used with CodeT5. Section \ref{Methodologies} describes the instructions used with Codex and ChatGPT. Section \ref{Results} reports the primary experimental results and observations. Finally in Section \ref{conclusion} we summarize the key conclusions and discuss possibilities for future work.

%% file: relatedwork.tex
Several recent works have utilized sequence to sequence models or pretrained Large Language Models for Code Generation tasks.  \citep{robustfill}, \citep{Rossol1986} are examples of papers which train a model specific to synthetic data generated on the basis of a formal grammar. Its not always feasible to parse an entire corpus in terms of formal grammar and use that for tokenization. However, a number of LLMs like CodeBERT \citep{CodeBERT}, CodeGPT \citep{CodeXGlue}, CodeT5 \citep{CodeT5}, are examples of models that have proven effective for python, C\#, Java etc. code generation with just a BPE tokenization. These models are often fine-tuned for a downstream task. Meanwhile the emergence of significantly larger models like Codex, GPT-3 etc have popularized the notion of in context learning \citep{fewshot}. In our work we explore both the modalities. 

While many operations research related work has been seen over the years, very few focuses on
auto-formulation of the optimization tasks from unstructured inputs such as text. Some of them
addresses partial formulation tasks and solver configuration. \citet{Iommazzo2020} discuss methods
to automatically configure solver parameters using Machine Learning. \citet{kiziltan2016} presents
methods to extract constraints from natural language. In the paper MathoptInterface \citet{mathoptint} suggests a standard representation for optimization problems compatible with variety of solvers.

In a recent paper associated with the NL4OPT competitions, \citet{NL4OPT} have
explained the use of transformers in extracting objective and constraints from natural language.
We setup auto formulation as Mathematical program synthesis from natural language input. 
Code or program generated by LLMs can suffer from several lexical, syntactic and semantic errors. Therefore a number of papers use grammar guidance for decoding ( \cite{picard}, \cite{jigsaw} ) or when suitable even use a symbolic or neural debugger for repairing the program ( \cite{REPL} ). 

Lack of data or parallel data is often a severe constraint in code generation and program synthesis tasks. Some of the approaches that have been adopted effectively include co-training,\citep{PLBART},  that allows a model to learn a join representation space for NL as well as Code. Some models employ back translation, \citep{Transcoder}, to improve the pretraining or increase the data augmentation for training. Recent works \citep{DataAug} and \citep{GPT3Mix}  have also explored the use of LLMs for generation of synthetic data which can be used for data augmentation.

%% file: data.tex
We use NL4OPT \citep{NL4OPT} dataset along with generated data to train our model and use only NL4OPT for evaluation. Data released as part of the NL4OPT competition consists three sets of word problems for linear programming from six different domains [Investment, Advertisement, Sales, Health Science, Transportation, Production]split into train(713), dev(99), and a private dataset test(290). Each linear word problem is accompanied by entity mapping and variable ordering. The labels are the Intermediate representation of the problem in either JSON format or XML which contains a list of constraints and a linear objective.

\lstset{
  basicstyle=\ttfamily,
  columns=fullflexible,
  frame=single,
  breaklines=true
}
\begin{lstlisting}[language=html, caption = Example sales optimization problem from the dataset ]
 A sports warehouse stocks rafts and kayaks. Each raft takes 
 10 sq ft of space while each kayak takes 12 sq ft of space. 
 The warehouse has 400 sq ft of space available. The warehouse
 has a budget of $10000 with each raft costing $200 and each 
 kayak costing $250. With rafting being much more popular in 
 the area, at least 55% of all items in stock must be rafts. 
 If the profit per raft is $45 and the profit per kayak 
 is $55, how many of each should be bought and sold 
 to maximize profit?
\end{lstlisting}

Let P\textsubscript{train}, P\textsubscript{dev} be the word problems in the train set and dev set respectively. Each problem in these sets has an objectives and 2 to 4 constraints . Please refer to the table \ref{tab:table1} below for different constraint types. 

\[\forall Obj: Obj\textsubscript{type} \in [linear,sum]\]
\[\forall Con: Con\textsubscript{type} \in (linear,sum,ratio,xy,xby,upperbound,lowerbound)\] 

\begin{table}[!ht]
    \centering
    \begin{tabular}{|l|l|}
    \hline
        \textbf{Constraint Type} & \textbf{Mathematical representation} \\ \hline
        sum &  $x + y \leq c$ \\ \hline
        upperbound & $x \leq c$ \\ \hline
        lowerbound & $x \geq c$  \\ \hline
        linear & $a\textsubscript{1} x + a\textsubscript{2} y \leq c$ \\ \hline
        ratio & $x \leq c (x+y)$ \\ \hline
        xby &  $x \leq a y$  \\ \hline
        xy &  x $\leq$ y  \\ \hline
    \end{tabular}
    \caption{Constraint Types used}
    \label{tab:table1}
\end{table}

\subsection{Evaluation Metrics}
\subsubsection{Canonical Accuracy - as in NL4OPT evaluation}
The evaluation is based on the correct declaration of objectives and constraints. Accuracy is calculated using the following formula:

\begin{equation}
    Canonical Accuracy = 1 - \frac{\sum(FP+FN)}{\sum D}
\end{equation}

where FP denotes the number of predicted objectives or constraints not matching with any of the actual declarations, FN denotes the number of actual declarations not matching with any of the predicted declarations and D denoted the total number of declarations in the ground truth.

\subsubsection{Execution Accuracy}
The above canonical metrics expect the language model to adhere to the same order of variable mappings as the labels. Meaning $a.X + b.Y \leq c$ is considered different from $b.Y + a.X \leq c$. To ensure the variable order is maintained an ordered variable mapping is expected as an input.

The fine-tuned models were able to adhere to the variable ordering if given as input, however, Codex, and ChatGPT were not able to adhere to the same and hallucinated variable names often. This made comparing canonical accuracy very difficult. Hence we propose using execution accuracy to compare fine tuned and GPT models.

To measure execution accuracy, we convert both actual and predicted declarations to python programs which solve for optimal values using Pulp \citep{Mitchell2011PuLPA}. We then compare the optimal values for exact match. This metric is used to compare program-only outputs such as codex.

%% file: setup.tex
We experimented with a variety of transformer based language models such as CodeT5, Bloom, Flan-T5 of various sizes. The common setup is to generate the intermediate representation in JSON format with the word problem as the input. One exception is OpenAI Codex \citep{codex}, which generated a python program using pulp library to formulate the problem. The performance of fine-tuned models were measured using the canonical accuracy of the IR generated and using execution match in the case of Codex.

\subsection{Fine tuning for Target and Auxillary tasks}
The models are trained for the target task which is to predict the intermediate representation of the problem in JSON format which then can be used to fill the program sketch created for linear optimization problems. We observed that out of the models we selected, CodeT5 \citep{CodeT5} trained on code generation tasks was performing better than others in our preliminary training (trained with only NL4OPT training data for 20 epochs). Hence we decided to take CodeT5 variants (small, base, large) as our base models for our experiments. Refer Appendix Table:\ref{table_prelim}

It has been proven that models trained on related subtasks such as scratchpads \citep{scratchpad}, and Chain of thought \citep{Chainofthought}, improve on the target tasks as well. We trained our models on the primary task which is to generate the declarations [objectives, constraints] as well as on sub-tasks such as (i) Predicting the number of constraints in the problem, (ii) Predicting the variable names,(iii) Predicting the parameter values,(iv) Direction of the objective.

All models were trained for 30 epochs on A100 40GB, with a learning rate of 5e-5, effective batch size of 32, and weight decay of 1e-5 with cross-entropy loss. The nl4opt data is provided with a entity tagging (output of a fine-tuned NER Model) used to enrich the text input. We train the models with both enriched and original data to minimize the dependency on the NER models.

\subsection{Augmentation using Language Models}
\begin{figure}[ht]
    \centering
    \includegraphics[width=15cm,height=12cm]{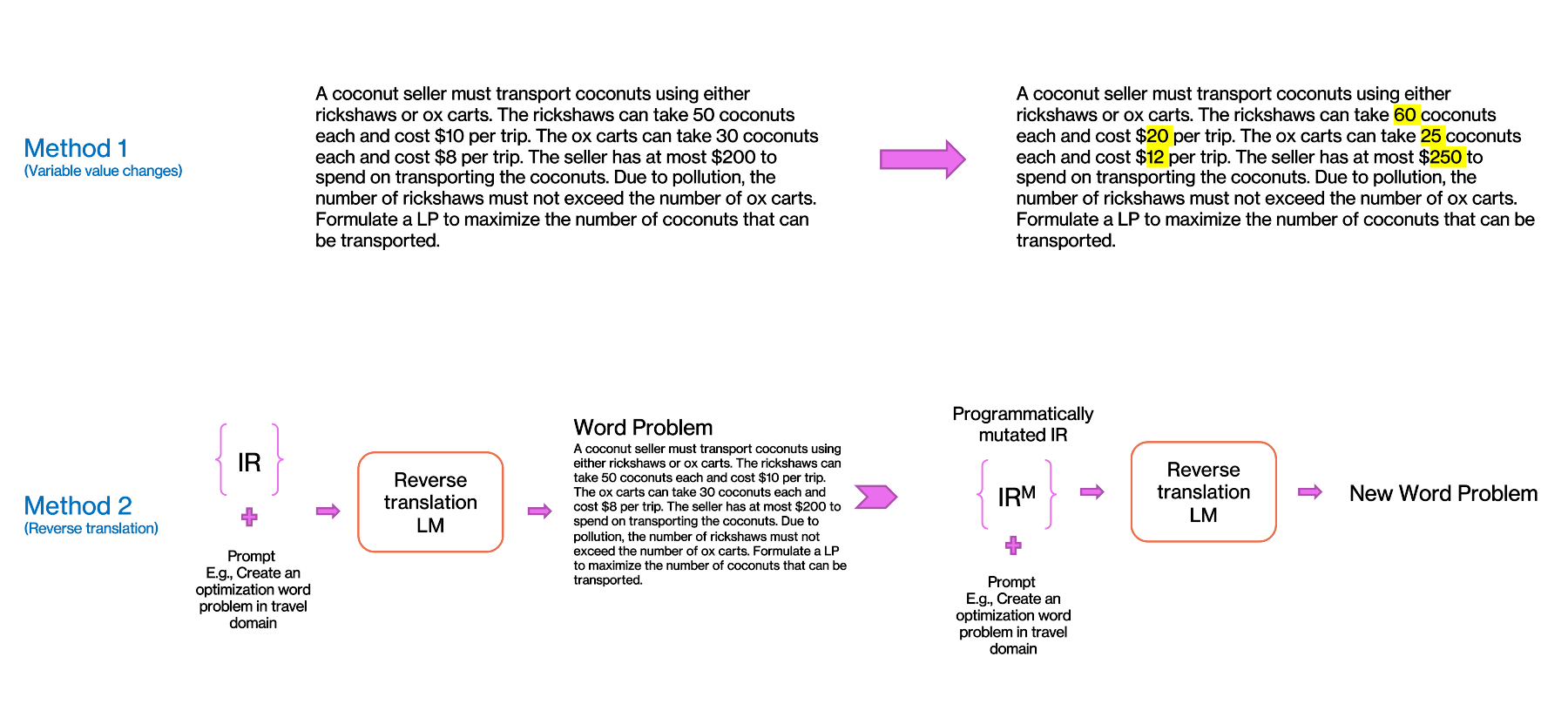}
    \caption{Data Augmentation methods}
    \label{fig:DataAug}
\end{figure}

The Natural Language for Optimization Problems (NL4OPT) dataset was the only available resource that featured optimization problems described in natural language. However, the dataset had several limitations, including the fact that it only contained single linear objective problems and was limited to a few (6)  industrial domains. Given that many real-world business problems are far more complex and involve multiple objectives, we decided to augment the dataset through parameter mutation and data synthesis utilizing the Generative Pre-trained Transformer 3 (GPT3) models. We observed that the generated word problems were not semantically correct, however, it introduced variations in text and domains so that the trained model performed better on the validation dataset. Refer \figureautorefname{\ref{fig:DataAug}}

The utilization of back translation through large language models, instead of templates, was motivated by the desire to simulate problems from various domains and to incorporate variations in the input text. The process involved transforming the intermediate representations from the NL4OPT dataset by mutating variables and parameter. The mutated intermediate representations were then utilized to generate human-readable problem descriptions through GPT3, which were validated and subsequently added to the training set. To imitate multi-objective problems, intermediate representations of complex single-objective problems with more than three linear constraints were selected and one of the linear constraints was converted into an additional objective simply by removing the limit.

The final augmented dataset used for training included both single-objective and multi-objective problems, whereas, the test data is originally from NL4OPT dataset which only consisted of single objective problem.

\begin{table}[htp]
    \centering
    \begin{tabular}{c|c|c|c}
        & Original & Simulated & Total\\
        \hline
    Single Objective & 713 & 3553 & 4226\\
    Multi Objective & 0 & 483 & 483 \\
    \hline
    Total &&& 4749
    \end{tabular}
    \caption{Augmented Training Data}
    \label{tab:my_label}
\end{table}

%% file: postprocess.tex
Optimization is prevalent across various industries and encompasses a diverse range of domain rules and knowledge, much of which may not be explicitly stated in the problem description but is nonetheless expected to be implicitly followed by the system. To address this, we employ symbolic post-processing to accurately select the appropriate sample from the generated outputs and to correct any known errors in the selected candidate. This approach enhances the reliability and robustness of the generated solutions. See in Appendix (figure \ref{fig:beamselection})

\subsubsection{Custom Beam Scorer}
Recent studies by \citet{hokamp-liu-2017-lexically} shows that constrained beam search improves accuracy of the generated text, We generate multiple samples using the models and along with the probability score from the model, used in combination execution guided scoring and rule based scoring. We observe accuracy improvements by applying few general semantics of Linear programming expressed as rules. These rules can be easily extended to include domain specific rules from knowledge bases or from domain text similar to the work by \citet{kiziltan2016}

To give an example of these rules, Let constraint Con(W,x,z) where W is the coefficient matrix, x is the variable and z is the RHS of the inequality. Then the based on simple linear equation properties, we know that:

\[\forall Con(W,x,z): con1 \neq con2 \land w1 \geq w2 \rightarrow z1 \geq z2\]
\[\forall Con(W,x,z): con1 \neq con2 \rightarrow w1 \neq w2\]

For each such rule violated, a penalty is applied on the corresponding beam, reducing the chance of that beam being selected. We then select 5 top candidates and calculate canonical accuracy to select the best one.  

\subsubsection{Automatic Post Editing using Domain Rules}
We perform certain corrections based on the domain rules, for example, if the objective function has the same coefficient for all the decision variables, it could be considered as a sum function rather than as linear.

\[ \forall Obj(W,x): w\textsubscript 1 == w\textsubscript 2 -> Obj\textsubscript type = 'objvar' \]

It was also beneficial to apply sanitizing such as duplicate constraint removal, removing terms when the constraint type is "sum" etc. Again, this rule set can be extended to have domain specific corrections and can operate completely decoupled from the model itself. This enables scaling of the model application to much wider industries.

\subsection{Configuring Optimization Solvers} \label{adapter}
\input{adapters}

%% file: adapters.tex
We created an adapter script to load the intermediate format (from the model) into an Optimization solver. We chose PULP as our optimization library since it had simple interfaces to configure the LP problem.

Solving for the optimized values enables newer evaluation/beam selection methods. The optimal values  from LLMs + Adapter can be compared with heuristics created by domain experts or generated from historical data. Using this method also enabled us to compare results with Codex GPT3 outputs since in Codex outputs, decision variable order was not strictly followed by the generated programs which made Canonical comparison difficult.

%% file: zeroshot.tex
 The intermediate representations were difficult for the LLMs such as Codex and ChatGPT even when prompted with semantically selected few shot examples. Hence, instead of prompting to generate the intermediate representation, we prompted Codex and ChatGPT to generate programs (refer \ref{codexprog}) that can solve the given optimization problem. The models tend to use a variety of optimization solvers (CVXPY, Pulp, Scipy). To maintain uniformity, We prompted to use Pulp as the default solver and a default variable name "prob" to hold the problem formulation (refer listing \ref{codexprog}). 

 ChatGPT (\url{https://chat.openai.com/}) being a conversational AI, generated additional explanations which were ignored and only generated code was extracted. Refer Appendix figure: \ref{fig:chatgpt1} and \ref{fig:chatgpt2}.

Using the adapter discussed in \ref{adapter} the actual declarations can be converted to similar python programs. The generated programs ( from actual and from LLMs ) when executed results in optimal values for the problem objective.  The results of the predicted and actual programs are then compared for execution matches.
 
\begin{lstlisting}[language=Python,  basicstyle=\tiny, caption=Example of program generated by Codex. Prompt text is enclosed in PROMPT tag for easy understanding, label=codexprog]
#<PROMPT>
"""
 A sports warehouse stocks rafts and kayaks. Each raft takes 
 10 sq ft of space while each kayak takes 12 sq ft of space. 
 The warehouse has 400 sq ft of space available. The warehouse
 has a budget of $10000 with each raft costing $200 and each 
 kayak costing $250. With rafting being much more popular in 
 the area, at least 55% of all items in stock must be rafts. 
 If the profit per raft is $45 and the profit per kayak 
 is $55, how many of each should be bought and sold 
 to maximize profit?
"""
# Use Pulp to solve the problem "prob_-1757358180.lp"
# Use the provided variable names rafts,kayaks
from pulp import *
# Create the "prob" variable to contain the problem data
problem_name = "prob_-1757358180"
prob = LpProblem(problem_name,
#<PROMPT>
                 LpMaximize)
# The 2 variables rafts and kayaks are created with a 
# lower limit of 0
rafts = LpVariable("rafts",
                   lowBound=0,
                   cat='Integer')  # @UndefinedVariable
kayaks = LpVariable("kayaks",
                    lowBound=0,
                    cat='Integer')  # @UndefinedVariable
# The objective function is added to "prob" first
prob += 45 * rafts + 55 * kayaks, "Profit"
# The two constraints are entered
prob += 10 * rafts + 12 * kayaks <= 400, "Space"
prob += 200 * rafts + 250 * kayaks <= 10000, "Cost"
prob += 0.55 * (rafts + kayaks) <= rafts, "Percentage"
# The problem data is written to an .lp file
prob.writeLP(problem_name + ".lp")
# The problem is solved using PuLP's choice of Solver
prob.solve()  # @UndefinedVariable
# The status of the solution is printed to the screen
print("Status:", LpStatus[prob.status])  # @UndefinedVariable
# Each of the variables is printed with 
# it's resolved optimum value
for v in prob.variables():  # @UndefinedVariable
    print(v.name, "=", v.varValue)
\end{lstlisting}

%% file: results.tex
We observed that codeT5-base model with custom beam scoring and correction performed better than other fine tuned models.   

To validate the new problems generated using back translation, we ensured that the parameters, decision variables, and objective/constraint directions were accurately represented in the text. However, we discovered that the word problems generated were not semantically correct in most cases. Hence, even though the models were trained on the augmented data, they were only evaluated using the original NL4OPT dataset. Our findings showed that training the models with the augmented data even when they are noisy, led to improved performance on the actual test data.

Ablation study shows that Subtask augmentation boosts the accuracy up by 10 percent in certain models. Similarly, training with noisy back translated data also boosts the accuracy by 9 points. Our rule based correction and scoring mechanism helps smaller models to achieve accuracy comparable with SOTA. We also show that with pass@k methods, we were able to surpass SOTA. 

\begin{table}[h]
    \centering
    \begin{tabular}{|l|l|l|l|}
    \hline
        \textbf{} & \textbf{code-t5-small} & \textbf{code-t5-base} & \textbf{code-t5-large} \\ \hline
        \textbf{No Augmentation} & 0.47 & 0.57 & 0.62 \\ \hline
        \textbf{Sub task Augmentation} & 0.6 & 0.69 & 0.72 \\ \hline
        \textbf{Back translation} & 0.69 & 0.78 & 0.78 \\ \hline
        \textbf{Sub Task + Back Translation} & 0.73 & 0.81 & 0.82 \\ \hline
        \textbf{Correction + Beam Search} & 0.7435 & 0.85 & 0.87 \\ \hline
        \textbf{Correction + Beam search (custom scoring) } & 0.74  & 0.8812 &  0.87 \\ \hline
    \end{tabular}
    \caption{Ablation Study on codeT5: Figure shows the accuracy levels reached by various models with a beam size of 5. We can observe that custom beam search helps the base model reach the accuracy similar to the larger ones.}
\end{table}
The state of the art model \citep{nl4optwinner} achieved accuracy of 0.882 from 5 beams on the dev set after 400 training epochs using BART large models without any data augmentation. We were able to achieve accuracy of 0.893 with pass@k (as defined in \cite{codex})  where k=5 beams with 30 training epochs by using data 
augmentation and post model symbolic methods. We see 3\% to 5\% increase in accuracy during inference time when logic guided corrections and custom beam search are used together. Further the percentage gain from the symbolic knowledge is more  for larger beam sizes ( refer Figure 3.), however after beam size 8, the accuracy started reducing.

\begin{table}[ht]
    \centering
    \begin{tabular}{|l|l|l|l|l|l|}
    \hline
        \textbf{Model Name} & \textbf{Size} & \textbf{Epochs} & \textbf{pass@k} & \textbf{Canonical Acc.} & \textbf{Method} \\ \hline
        \textbf{Baseline} & 140M & 200 & - & 63 & Prompt Guided \\ \hline
        \textbf{Current SOTA} & 400M & 400 & 5 & 88.2 & All at once \\ \hline
        \textbf{Proposed Model} & 220M & 30 & 5 & \textbf{89.63} & Logic guided Correction\\ \hline
    \end{tabular}
    \caption{Comparison of Baseline (BART), Current SOTA (Bart Large) \cite{nl4optwinner}, Proposed (CodeT5 + Logic Guided Correction) performances }
\end{table}

We also compared the best model with Codex and ChatGPT outputs on exact match of optimal values since there is no order mapping of variables possible for codex outputs which makes canonical comparison difficult. We can see that the fine tuned model with logic guided scoring and correction was able to outperform Codex. We can infer that learning the intermediate representation enables models to more accurately solve the problem.  Codex was able to score 36\% on exact match, ChatGPT was able to score 41\%, whereas our finetuned models were able to achieve 73\% exact match. 

\begin{table}[!ht]
    \centering
    \begin{tabular}{|l|l|l|l|}
    \hline
         & \textbf{codeT5} & \textbf{Codex} & \textbf{ChatGPT} \\ \hline
        \textbf{Execution Match} & 0.73 & 0.36 & 0.41 \\ \hline
    \end{tabular}
    \caption{Execution match is calculated by executing the generated program to find the optimal objective value and finding exact match against actual}
\end{table}

%% file: conclusion.tex
In this paper, we considered the task of formulating optimization problems from natural language. The dataset we considered had domain-specific variations, and thus with only 730 training samples, the data can be considered to be sparse. We evaluated several models and proposed methods that enable the use of large language models to synthesize the canonical forms as well as python code despite the task being a low-data task. We observed progressive utility for the use of auxiliary tasks, GPT-3 based data augmentation, and logic based post-processing of decoding beams. Since we extended the task to code synthesis, as against just classification as done in the NL4OPT challenge, we were able to measure the efficacy of the methods in terms of execution accuracy.

%% file: Appendix.tex
\subsection{Initial Model selection}
We selected CodeT5 for our experiment since we setup our problem as a program synthesis task and it performed better compared to other similar sized models. 

\begin{table}[!ht]
    \centering
    \begin{tabular}{|l|l|l|l|l|l|}
    \hline
        \textbf{Model Name} & \textbf{Inputs} & \textbf{Output} & \textbf{Canonical accuracy} & \textbf{Rouge} & \textbf{Epochs} \\ \hline
        \textbf{code-t5-small} & Doc & JSON & 47\% & 83\% & 30 \\ \hline
        \textbf{code-t5-small} & Doc + Vars + Ord & JSON & 52\% & 82\% & 30 \\ \hline
        \textbf{code-t5-base} & Doc & JSON & 57\% & 83\% & 20 \\ \hline
        \textbf{code-t5-base} & Doc + Vars + Ord & JSON & 64\% & 84\% & 20 \\ \hline
        \textbf{code-t5-large} & Doc & JSON & 67\% & NA & 10 \\ \hline
        \textbf{flan-t5-large} & Doc & JSON & 27\% & 79\% & 10 \\ \hline
        \textbf{gpt3-curie} & Doc & JSON & 17\% & NA & 4 \\ \hline
    \end{tabular}
    \caption{All models trained on unaugmented data, small models for 30 epochs, base ofr 20 and large ones for 10 epochs. The type of input specifies what texts were concatenated from the input records. }
    \label{table_prelim}
\end{table}

\subsection{Beam size vs Canonical Accuracy}
We see that with logic guided corrections, larger beam sizes until a certain size gain accuracy compared to its non guided pair.

\begin{figure}[h]
    \centering
    \includegraphics[scale=0.75]{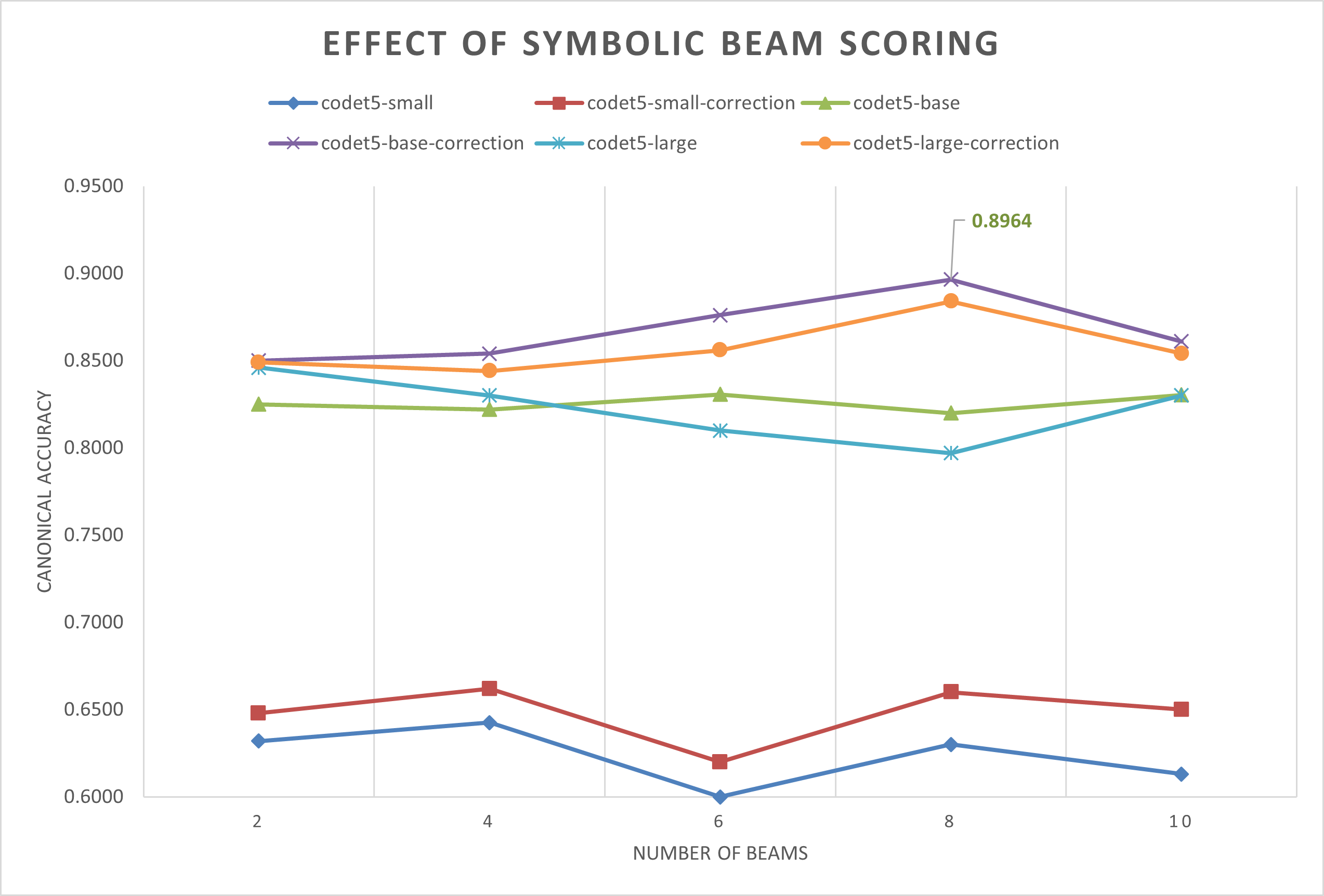}
    \caption{Effect of custom beam search on Accuracy}
    \label{fig:Figure 4}
\end{figure}

\subsection{Error Analysis}
We will discuss frequently seen errors in this section, we believe that most of them can be addressed with carefully designed rules and data augmentation. 

\textbf{Missing variables in the objective.}
\begin{lstlisting}[language=Python]
Actual : {'obj_declaration': {'type': 'objective', 'direction': 'maximize', 'name': 'viewers', 'terms': {'magazine ad': '10000', 'flyer': '5000', 'billboard ad': '25000'}}
Predicted: {'obj_declaration': {'type': 'objective', 'direction': 'maximize', 'name': 'viewers', 'terms': {'flyer': '5000', 'billboard ad': '25000'}}
\end{lstlisting}

\textbf{Correct Prediction, however label is missing a constraint}: As we could see the constraint that a train could atmost carry 500 passangers is not in the original label. Similarly we found 2 more cases where predicted actually was correct.

\begin{lstlisting}
text:'A train can carry at most 500 passengers. It has two seat types: AC seats, and non-AC seats (AC is air conditioned). A profit of $50 is made on each AC seat ticket and a profit of $30 is made on each non-AC seat ticket. The train company reserves at least 100 seats as AC seats. However, a minimum of 2 times as many passengers prefer to travel on non-AC seats than on AC seats. How many seat tickets of each type should be sold to maximize profit? What is that profit?'

Actual:
{'obj_declaration': {'type': 'objective', 'direction': 'maximize', 'name': 'profit', 'terms': {'AC seat': '50', 'non-AC seat': '30'}}, 'const_declarations': [{'type': 'lowerbound', 'direction': 'at least', 'limit': '100', 'var': 'AC seats', 'operator': 'GREATER_OR_EQUAL'}, {'type': 'xby', 'x_var': 'non-AC seats', 'direction': 'minimum', 'param': '2', 'y_var': 'AC seats', 'operator': 'GREATER_OR_EQUAL'}], 'vars': ['AC seats', 'non-AC seats']}
---------------------------------------
Predicted
{'obj_declaration': {'type': 'objective', 'direction': 'maximize', 'name': 'profit', 'terms': {'AC seat': '50', 'non-AC seat': '30'}}, 'const_declarations': [{'type': 'sum', 'direction': 'at most', 'limit': '500', 'operator': 'LESS_OR_EQUAL'}, {'type': 'lowerbound', 'direction': 'at least', 'limit': '100', 'var': 'AC seats', 'operator': 'GREATER_OR_EQUAL'}, {'type': 'xby', 'x_var': 'non-AC seats', 'direction': 'minimum', 'param': '2', 'y_var': 'AC seats', 'operator': 'GREATER_OR_EQUAL'}], 'vars': ['AC seats', 'non-AC seats'], 'id': '-996226930'}
\end{lstlisting}

\textbf{Missing Constraint}:In few cases, the predicted missed generating a constraint all together in all the beams generated.

\begin{lstlisting}
Actual:
{'obj_declaration': {'type': 'objvar', 'direction': 'minimize', 'name': 'total number of stores', 'vars': ['film-based', 'electrical-based stores']}, 'const_declarations': [{'type': 'xby', 'x_var': 'electrical-based stores', 'direction': 'at least', 'param': 'two', 'y_var': 'film-based stores', 'operator': 'GREATER_OR_EQUAL'}, {'type': 'lowerbound', 'direction': 'at least', 'limit': '5', 'var': 'film-based stores', 'operator': 'GREATER_OR_EQUAL'}, {'type': 'linear', 'direction': 'at least', 'limit': '170', 'terms': {'Film-based stores': '2', 'electrical-based store': 'four'}, 'operator': 'GREATER_OR_EQUAL'}, {'type': 'linear', 'direction': 'at most', 'limit': '600', 'terms': {'electric-based stores': '15', 'Film-based stores': '10'}, 'operator': 'LESS_OR_EQUAL'}], 'vars': ['film-based', 'electrical-based stores']}
----------------------------------------
Predicted
{'obj_declaration': {'type': 'objvar', 'direction': 'minimize', 'name': 'total number of stores', 'vars': ['film-based', 'electrical-based stores']}, 'const_declarations': [{'type': 'lowerbound', 'direction': 'at least', 'limit': '5', 'var': 'film-based stores', 'operator': 'GREATER_OR_EQUAL'}, {'type': 'linear', 'direction': 'at least', 'limit': '170', 'terms': {'Film-based stores': '2', 'electrical-based store': 'four'}, 'operator': 'GREATER_OR_EQUAL'}, {'type': 'linear', 'direction': 'at most', 'limit': '600', 'terms': {'Film-based stores': '10', 'electric-based stores': '15'}, 'operator': 'LESS_OR_EQUAL'}], 'vars': ['film-based', 'electrical-based stores'], 'id': '-1194187124'}
\end{lstlisting}

\textbf{Swapped limits:} The limits of the constraints gets swapped between two constraints.

\begin{lstlisting}
Actual:
{'obj_declaration': {'type': 'objvar', 'direction': 'minimize', 'name': 'total number of shifts', 'vars': ['dentists', 'oral hygienists']}, 'const_declarations': [{'type': 'linear', 'direction': 'require', 'limit': '1000', 'terms': {'Dentists': '12', 'oral hygienists': '5'}, 'operator': 'GREATER_OR_EQUAL'}, {'type': 'sum', 'direction': 'at least', 'limit': '20', 'operator': 'GREATER_OR_EQUAL'}, {'type': 'linear', 'direction': 'budget', 'limit': '65000', 'terms': {'dentists': '900', 'oral hygienists': '250'}, 'operator': 'LESS_OR_EQUAL'}], 'vars': ['dentists', 'oral hygienists']}
------------------------------------------------
Predicted
{'obj_declaration': {'type': 'objvar', 'direction': 'minimize', 'name': 'total number of shifts', 'vars': ['dentists', 'oral hygienists']}, 'const_declarations': [{'type': 'linear', 'direction': 'budget', 'limit': '65000', 'terms': {'Dentists': '12', 'oral hygienists': '5'}, 'operator': 'LESS_OR_EQUAL'}, {'type': 'sum', 'direction': 'at least', 'limit': '20', 'operator': 'GREATER_OR_EQUAL'}, {'type': 'linear', 'direction': 'require', 'limit': '1000', 'terms': {'Dentists': '900', 'oral hygienists': '250'}, 'operator': 'GREATER_OR_EQUAL'}], 'vars': ['dentists', 'oral hygienists']}
\end{lstlisting}
\textbf{Syntax Errors} Model hallucinates and produces incorrect literals which makes that beam invalid even though it has the right formulation;
\begin{lstlisting}
num_of_constraints = 2
obj_declaration = {'type': 'objvar', 'direction':'minimize', 'name': 'number of branches', 'vars': ['urban branches','remote branch']}
const_declarations = [{'type': 'linear', 'direction': 'available', 'limit': '550', 'terms': {'remote branch': '5', 'urban branch': '12'}, 'operator': 'LESS_OR_EQUAL'}, {'type': 'linear', 'direction': 'at least', 'limit': '1200', 'terms': {'remote branch': '2', 'urban branch': '7'}, 'operator': 'GREATER_OR_EQUAL'}]
 yield'}]
\end{lstlisting}
\newpage
\subsection{Post Processing Rules - subset}
\textbf{Post model Corrections:}
\begin{lstlisting}

1. If a Linear objective's terms have same coefficients, convert the objective into a 'objvar' type
2. If the same coefficients are being reused in the constraints, most probably it is a 'objvar' type.
3. If the constraint type is "lowerbound" the operator should be "GREATER_THAN_EQUAL_TO" and vice versa
4. if the constraint type is "sum" then there should be no terms allowed for the variables (i.e coeff of all variables is 1)
\end{lstlisting}

\textbf{Beam Search scoring:}
\begin{lstlisting}
Violation of each rule reduces the corresponding beam's score:

1. There should be at least as many constraints as the constraint directions available in the SPANs from tagged input.
2. All the limits ( RHS of the inequality ) must be available in the formulation, proportionate penalty  will be imposed on the total score
3. All the parameters ( coefficients in LHS ) should be available in the formulation, penalty will be imposed proportionately.
4. The scale difference of the coefficients for a given constraint has to be below the maximum threshold. Proportionate penalty to be applied on the total score.
5. If there is a postive constraint on the decision variables, the constraints must be positive as well. Meaning increasing the parameter value should also increase the limit.
\end{lstlisting}

\begin{figure}[h]
    \centering
    \includegraphics[scale=0.45]{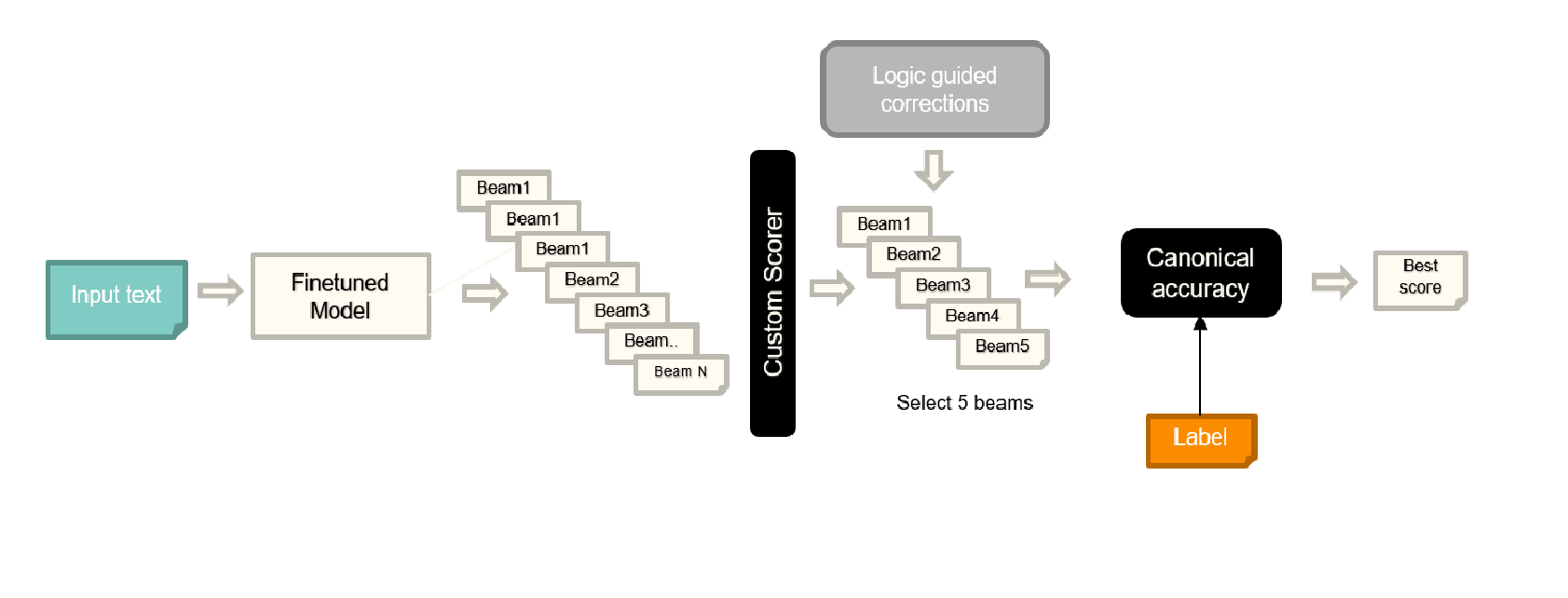}
    \caption{Candidate selection process}
    \label{fig:beamselection}
\end{figure}

\subsection{ChatGPT Experiments}
\begin{figure}[htp]
    \centering
    \includegraphics[scale=0.75]{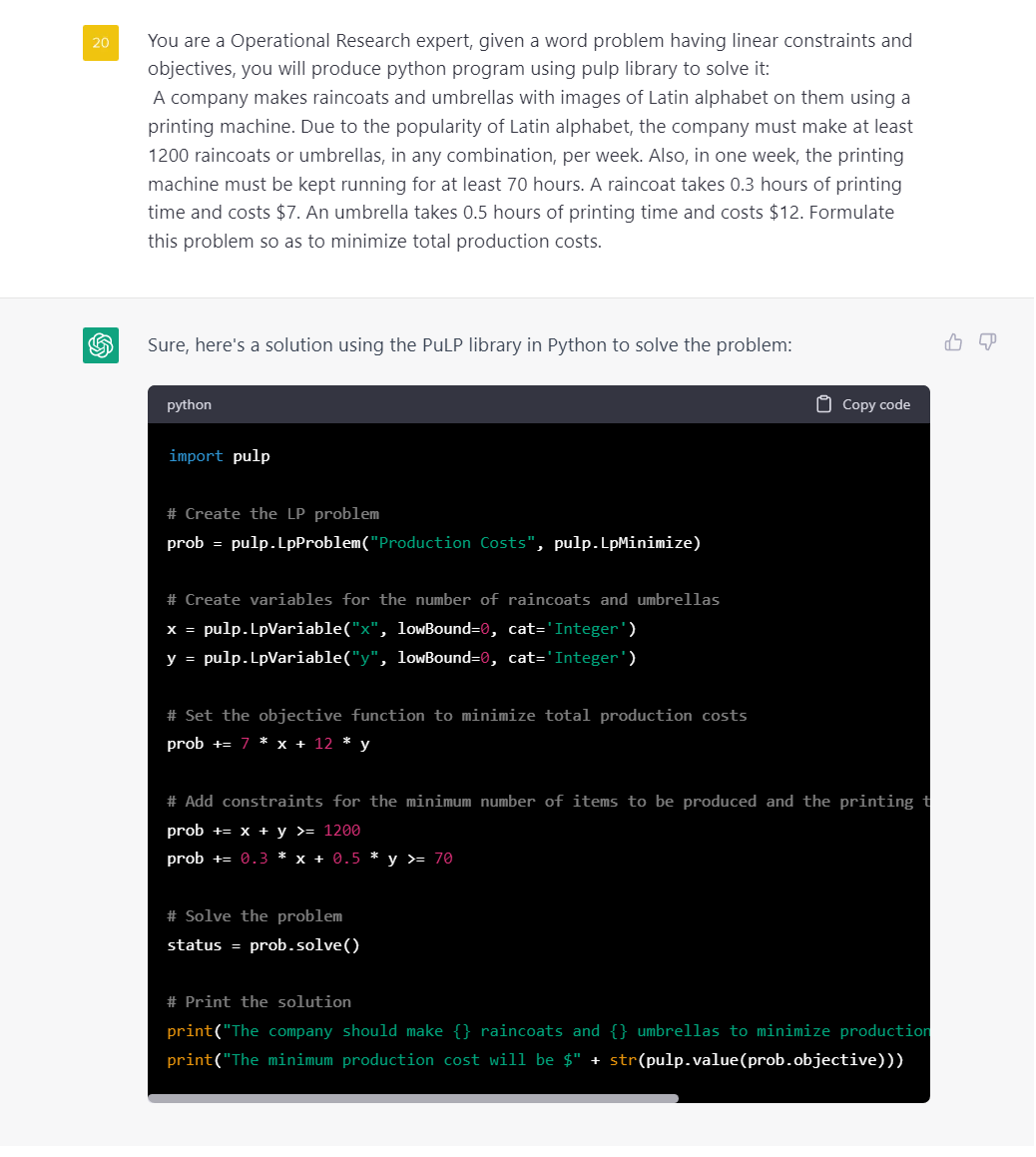}
    \caption{ChatGPT - Question 1}
    \label{fig:chatgpt1}
\end{figure}

\begin{figure}[htp]
    \centering
    \includegraphics[scale=0.75]{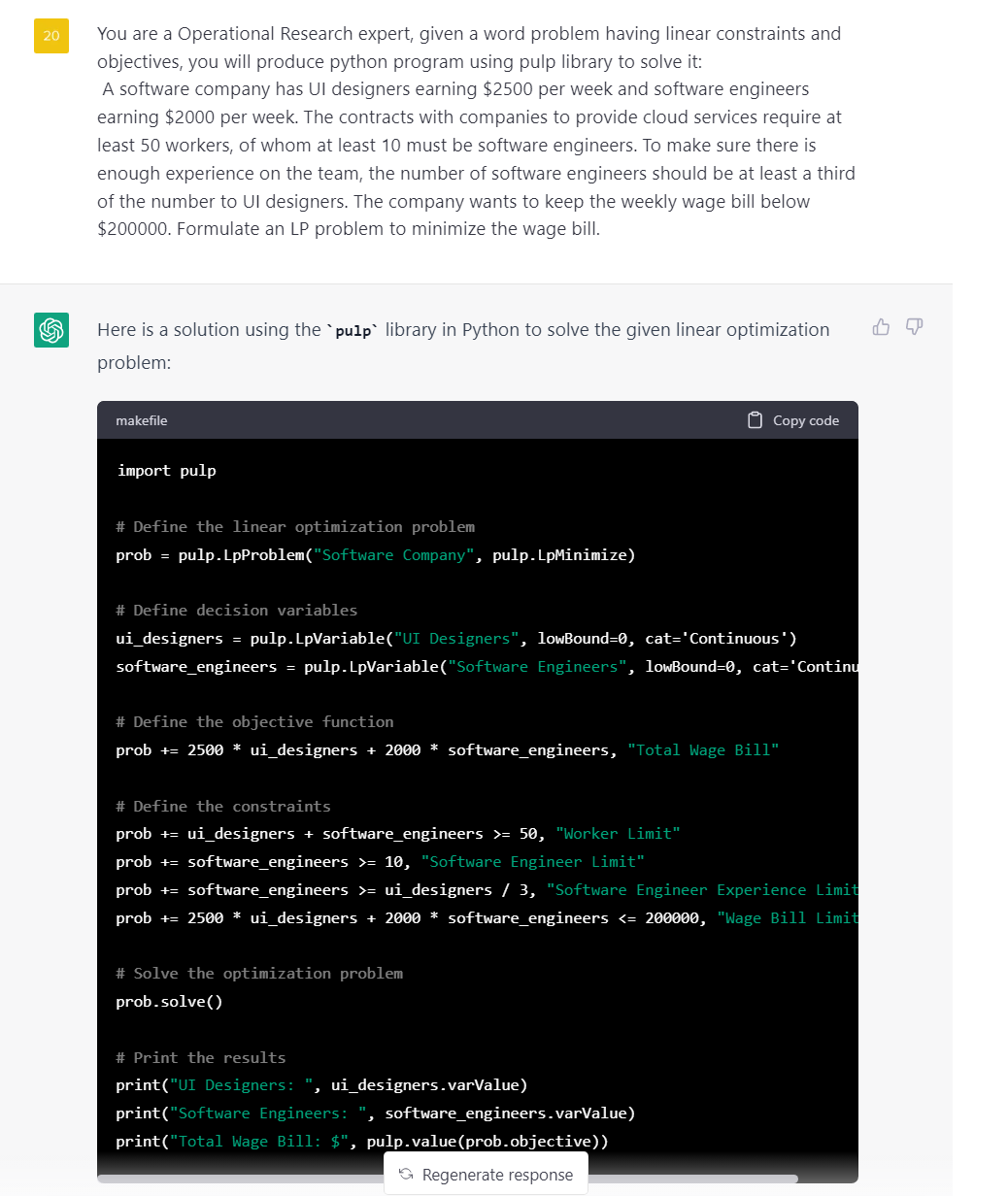}
    \caption{ChatGPT - Question 2}
    \label{fig:chatgpt2}
\end{figure}

Since official API for ChatGPT was not available at the time of writing this paper, we had to rely on manual methods to prompt ChatGPT. We prompted with a standard prompt for all the problems. Copied the response from ChatGPT and run it without any modification. Refer Figure \ref{fig:chatgpt1} and \ref{fig:chatgpt2}. We found few examples where the use of pulp variables were wrong and resulted in a Exception. Such examples where marked with a 0 for optimal value. 


%% file: main.bbl
\begin{thebibliography}{43}
\providecommand{\natexlab}[1]{#1}
\providecommand{\url}[1]{\texttt{#1}}
\expandafter\ifx\csname urlstyle\endcsname\relax
  \providecommand{\doi}[1]{doi: #1}\else
  \providecommand{\doi}{doi: \begingroup \urlstyle{rm}\Url}\fi

\bibitem[Ahmad et~al.(2021)Ahmad, Chakraborty, Ray, and Chang]{PLBART}
Wasi~Uddin Ahmad, Saikat Chakraborty, Baishakhi Ray, and Kai-Wei Chang.
\newblock Unified pre-training for program understanding and generation, 2021.
\newblock URL \url{https://arxiv.org/abs/2103.06333}.

\bibitem[Bengio \& LeCun(2007)Bengio and LeCun]{Bengio+chapter2007}
Yoshua Bengio and Yann LeCun.
\newblock Scaling learning algorithms towards {AI}.
\newblock In \emph{Large Scale Kernel Machines}. MIT Press, 2007.

\bibitem[Brown et~al.(2020)Brown, Mann, Ryder, Subbiah, Kaplan, Dhariwal,
  Neelakantan, Shyam, Sastry, Askell, Agarwal, Herbert-Voss, Krueger, Henighan,
  Child, Ramesh, Ziegler, Wu, Winter, Hesse, Chen, Sigler, Litwin, Gray, Chess,
  Clark, Berner, McCandlish, Radford, Sutskever, and Amodei]{fewshot}
Tom~B. Brown, Benjamin Mann, Nick Ryder, Melanie Subbiah, Jared Kaplan,
  Prafulla Dhariwal, Arvind Neelakantan, Pranav Shyam, Girish Sastry, Amanda
  Askell, Sandhini Agarwal, Ariel Herbert-Voss, Gretchen Krueger, Tom Henighan,
  Rewon Child, Aditya Ramesh, Daniel~M. Ziegler, Jeffrey Wu, Clemens Winter,
  Christopher Hesse, Mark Chen, Eric Sigler, Mateusz Litwin, Scott Gray,
  Benjamin Chess, Jack Clark, Christopher Berner, Sam McCandlish, Alec Radford,
  Ilya Sutskever, and Dario Amodei.
\newblock Language models are few-shot learners, 2020.
\newblock URL \url{https://arxiv.org/abs/2005.14165}.

\bibitem[Cambazard et~al.(2021)Cambazard, Catusse, Brauner, and
  Lemaire]{Cambazard2021}
Hadrien Cambazard, Nicolas Catusse, Nadia Brauner, and Pierre Lemaire.
\newblock Teaching {OR}: automatic evaluation for linear programming modelling.
\newblock \emph{4OR}, 20\penalty0 (2):\penalty0 333--345, July 2021.
\newblock \doi{10.1007/s10288-021-00488-9}.
\newblock URL \url{https://doi.org/10.1007/s10288-021-00488-9}.

\bibitem[Chen et~al.(2021)Chen, Tworek, Jun, Yuan, Pinto, Kaplan, Edwards,
  Burda, Joseph, Brockman, Ray, Puri, Krueger, Petrov, Khlaaf, Sastry, Mishkin,
  Chan, Gray, Ryder, Pavlov, Power, Kaiser, Bavarian, Winter, Tillet, Such,
  Cummings, Plappert, Chantzis, Barnes, Herbert-Voss, Guss, Nichol, Paino,
  Tezak, Tang, Babuschkin, Balaji, Jain, Saunders, Hesse, Carr, Leike, Achiam,
  Misra, Morikawa, Radford, Knight, Brundage, Murati, Mayer, Welinder, McGrew,
  Amodei, McCandlish, Sutskever, and Zaremba]{codex}
Mark Chen, Jerry Tworek, Heewoo Jun, Qiming Yuan, Henrique Ponde de~Oliveira
  Pinto, Jared Kaplan, Harri Edwards, Yuri Burda, Nicholas Joseph, Greg
  Brockman, Alex Ray, Raul Puri, Gretchen Krueger, Michael Petrov, Heidy
  Khlaaf, Girish Sastry, Pamela Mishkin, Brooke Chan, Scott Gray, Nick Ryder,
  Mikhail Pavlov, Alethea Power, Lukasz Kaiser, Mohammad Bavarian, Clemens
  Winter, Philippe Tillet, Felipe~Petroski Such, Dave Cummings, Matthias
  Plappert, Fotios Chantzis, Elizabeth Barnes, Ariel Herbert-Voss,
  William~Hebgen Guss, Alex Nichol, Alex Paino, Nikolas Tezak, Jie Tang, Igor
  Babuschkin, Suchir Balaji, Shantanu Jain, William Saunders, Christopher
  Hesse, Andrew~N. Carr, Jan Leike, Josh Achiam, Vedant Misra, Evan Morikawa,
  Alec Radford, Matthew Knight, Miles Brundage, Mira Murati, Katie Mayer, Peter
  Welinder, Bob McGrew, Dario Amodei, Sam McCandlish, Ilya Sutskever, and
  Wojciech Zaremba.
\newblock Evaluating large language models trained on code, 2021.
\newblock URL \url{https://arxiv.org/abs/2107.03374}.

\bibitem[Chung et~al.(2022)Chung, Hou, Longpre, Zoph, Tay, Fedus, Li, Wang,
  Dehghani, Brahma, Webson, Gu, Dai, Suzgun, Chen, Chowdhery, Castro-Ros,
  Pellat, Robinson, Valter, Narang, Mishra, Yu, Zhao, Huang, Dai, Yu, Petrov,
  Chi, Dean, Devlin, Roberts, Zhou, Le, and Wei]{flant5}
Hyung~Won Chung, Le~Hou, Shayne Longpre, Barret Zoph, Yi~Tay, William Fedus,
  Yunxuan Li, Xuezhi Wang, Mostafa Dehghani, Siddhartha Brahma, Albert Webson,
  Shixiang~Shane Gu, Zhuyun Dai, Mirac Suzgun, Xinyun Chen, Aakanksha
  Chowdhery, Alex Castro-Ros, Marie Pellat, Kevin Robinson, Dasha Valter,
  Sharan Narang, Gaurav Mishra, Adams Yu, Vincent Zhao, Yanping Huang, Andrew
  Dai, Hongkun Yu, Slav Petrov, Ed~H. Chi, Jeff Dean, Jacob Devlin, Adam
  Roberts, Denny Zhou, Quoc~V. Le, and Jason Wei.
\newblock Scaling instruction-finetuned language models, 2022.
\newblock URL \url{https://arxiv.org/abs/2210.11416}.

\bibitem[Devlin et~al.(2017{\natexlab{a}})Devlin, Bunel, Singh, Hausknecht, and
  Kohli]{karelio}
Jacob Devlin, Rudy Bunel, Rishabh Singh, Matthew Hausknecht, and Pushmeet
  Kohli.
\newblock Neural program meta-induction, 2017{\natexlab{a}}.
\newblock URL \url{https://arxiv.org/abs/1710.04157}.

\bibitem[Devlin et~al.(2017{\natexlab{b}})Devlin, Uesato, Bhupatiraju, Singh,
  Mohamed, and Kohli]{robustfill}
Jacob Devlin, Jonathan Uesato, Surya Bhupatiraju, Rishabh Singh, Abdel-rahman
  Mohamed, and Pushmeet Kohli.
\newblock Robustfill: Neural program learning under noisy i/o,
  2017{\natexlab{b}}.
\newblock URL \url{https://arxiv.org/abs/1703.07469}.

\bibitem[Diamond \& Boyd(2016)Diamond and Boyd]{CVXPY}
Steven Diamond and Stephen Boyd.
\newblock Cvxpy: A python-embedded modeling language for convex optimization,
  2016.
\newblock URL \url{https://arxiv.org/abs/1603.00943}.

\bibitem[Elizarov et~al.(2014)Elizarov, Kirillovich, Lipachev, Nevzorova,
  Solovyev, and Zhiltsov]{https://doi.org/10.48550/arxiv.1408.6806}
Alexander Elizarov, Alexander Kirillovich, Evgeny Lipachev, Olga Nevzorova,
  Valery Solovyev, and Nikita Zhiltsov.
\newblock Mathematical knowledge representation: Semantic models and
  formalisms, 2014.
\newblock URL \url{https://arxiv.org/abs/1408.6806}.

\bibitem[Ellis et~al.(2019)Ellis, Nye, Pu, Sosa, Tenenbaum, and
  Solar-Lezama]{REPL}
Kevin Ellis, Maxwell Nye, Yewen Pu, Felix Sosa, Josh Tenenbaum, and Armando
  Solar-Lezama.
\newblock Write, execute, assess: Program synthesis with a repl, 2019.
\newblock URL \url{https://arxiv.org/abs/1906.04604}.

\bibitem[Feng et~al.(2020)Feng, Guo, Tang, Duan, Feng, Gong, Shou, Qin, Liu,
  Jiang, and Zhou]{CodeBERT}
Zhangyin Feng, Daya Guo, Duyu Tang, Nan Duan, Xiaocheng Feng, Ming Gong, Linjun
  Shou, Bing Qin, Ting Liu, Daxin Jiang, and Ming Zhou.
\newblock Codebert: A pre-trained model for programming and natural languages,
  2020.
\newblock URL \url{https://arxiv.org/abs/2002.08155}.

\bibitem[Gangwar \& Kani(2022)Gangwar and Kani]{nl4optwinner}
Neeraj Gangwar and Nickvash Kani.
\newblock Highlighting named entities in input for auto-formulation of
  optimization problems, 2022.
\newblock URL \url{https://arxiv.org/abs/2212.13201}.

\bibitem[Han et~al.(2022)Han, Schoelkopf, Zhao, Qi, Riddell, Benson, Sun,
  Zubova, Qiao, Burtell, Peng, Fan, Liu, Wong, Sailor, Ni, Nan, Kasai, Yu,
  Zhang, Joty, Fabbri, Kryscinski, Lin, Xiong, and Radev]{FOLIO}
Simeng Han, Hailey Schoelkopf, Yilun Zhao, Zhenting Qi, Martin Riddell, Luke
  Benson, Lucy Sun, Ekaterina Zubova, Yujie Qiao, Matthew Burtell, David Peng,
  Jonathan Fan, Yixin Liu, Brian Wong, Malcolm Sailor, Ansong Ni, Linyong Nan,
  Jungo Kasai, Tao Yu, Rui Zhang, Shafiq Joty, Alexander~R. Fabbri, Wojciech
  Kryscinski, Xi~Victoria Lin, Caiming Xiong, and Dragomir Radev.
\newblock Folio: Natural language reasoning with first-order logic, 2022.
\newblock URL \url{https://arxiv.org/abs/2209.00840}.

\bibitem[Hentenryck(1999)]{Hentenryck1999TheOO}
Pascal~Van Hentenryck.
\newblock The opl optimization programming language.
\newblock In \emph{The OPL optimization programming language}, 1999.

\bibitem[Hinton et~al.(2006)Hinton, Osindero, and Teh]{Hinton06}
Geoffrey~E. Hinton, Simon Osindero, and Yee~Whye Teh.
\newblock A fast learning algorithm for deep belief nets.
\newblock \emph{Neural Computation}, 18:\penalty0 1527--1554, 2006.

\bibitem[Hokamp \& Liu(2017)Hokamp and Liu]{hokamp-liu-2017-lexically}
Chris Hokamp and Qun Liu.
\newblock Lexically constrained decoding for sequence generation using grid
  beam search.
\newblock In \emph{Lexically Constrained Decoding for Sequence Generation Using
  Grid Beam Search}. Association for Computational Linguistics, 2017.
\newblock \doi{10.18653/v1/P17-1141}.
\newblock URL \url{https://aclanthology.org/P17-1141}.

\bibitem[Huang et~al.(2021)Huang, Wu, Jeong, Wang, Chen, and Hwu]{pylog}
Sitao Huang, Kun Wu, Hyunmin Jeong, Chengyue Wang, Deming Chen, and Wen-Mei
  Hwu.
\newblock Pylog: An algorithm-centric python-based fpga programming and
  synthesis flow.
\newblock \emph{IEEE Transactions on Computers}, 70\penalty0 (12):\penalty0
  2015--2028, 2021.
\newblock \doi{10.1109/TC.2021.3123465}.

\bibitem[Iommazzo et~al.(2020)Iommazzo, D'Ambrosio, Frangioni, and
  Liberti]{Iommazzo2020}
Gabriele Iommazzo, Claudia D'Ambrosio, Antonio Frangioni, and Leo Liberti.
\newblock A learning-based mathematical programming formulation for the
  automatic configuration of optimization solvers.
\newblock In \emph{Machine Learning, Optimization, and Data Science}, pp.\
  700--712. Springer International Publishing, 2020.
\newblock \doi{10.1007/978-3-030-64583-0_61}.
\newblock URL \url{https://doi.org/10.1007/978-3-030-64583-0_61}.

\bibitem[Jain et~al.(2021)Jain, Vaidyanath, Iyer, Natarajan, Parthasarathy,
  Rajamani, and Sharma]{jigsaw}
Naman Jain, Skanda Vaidyanath, Arun Iyer, Nagarajan Natarajan, Suresh
  Parthasarathy, Sriram Rajamani, and Rahul Sharma.
\newblock Jigsaw: Large language models meet program synthesis, 2021.
\newblock URL \url{https://arxiv.org/abs/2112.02969}.

\bibitem[KIMMIG et~al.(2011)KIMMIG, DEMOEN, RAEDT, COSTA, and
  ROCHA]{KIMMIG_2011}
ANGELIKA KIMMIG, BART DEMOEN, LUC~DE RAEDT, V{\'{I} }TOR~SANTOS COSTA, and
  RICARDO ROCHA.
\newblock On the implementation of the probabilistic logic programming language
  {ProbLog}.
\newblock \emph{Theory and Practice of Logic Programming}, 11\penalty0
  (2-3):\penalty0 235--262, jan 2011.
\newblock \doi{10.1017/s1471068410000566}.
\newblock URL \url{https://doi.org/10.1017%2Fs1471068410000566}.

\bibitem[Kiziltan et~al.(2016{\natexlab{a}})Kiziltan, Lippi, and
  Torroni]{10.5555/3060621.3060725}
Zeynep Kiziltan, Marco Lippi, and Paolo Torroni.
\newblock Constraint detection in natural language problem descriptions.
\newblock In \emph{Proceedings of the Twenty-Fifth International Joint
  Conference on Artificial Intelligence}, IJCAI'16, pp.\  744–750. AAAI
  Press, 2016{\natexlab{a}}.
\newblock ISBN 9781577357704.

\bibitem[Kiziltan et~al.(2016{\natexlab{b}})Kiziltan, Lippi, and
  Torroni]{kiziltan2016}
Zeynep Kiziltan, Marco Lippi, and Paolo Torroni.
\newblock Constraint detection in natural language problem descriptions.
\newblock \emph{Neural Computation}, 18:\penalty0 1527--1554,
  2016{\natexlab{b}}.

\bibitem[Lachaux et~al.(2020)Lachaux, Roziere, Chanussot, and
  Lample]{Transcoder}
Marie-Anne Lachaux, Baptiste Roziere, Lowik Chanussot, and Guillaume Lample.
\newblock Unsupervised translation of programming languages, 2020.
\newblock URL \url{https://arxiv.org/abs/2006.03511}.

\bibitem[Legat et~al.(2020)Legat, Dowson, Garcia, and Lubin]{mathoptint}
Benoit Legat, Oscar Dowson, Joaquim~Dias Garcia, and Miles Lubin.
\newblock Mathoptinterface: a data structure for mathematical optimization
  problems, 2020.
\newblock URL \url{https://arxiv.org/abs/2002.03447}.

\bibitem[Lu et~al.(2021)Lu, Guo, Ren, Huang, Svyatkovskiy, Blanco, Clement,
  Drain, Jiang, Tang, Li, Zhou, Shou, Zhou, Tufano, Gong, Zhou, Duan,
  Sundaresan, Deng, Fu, and Liu]{CodeXGlue}
Shuai Lu, Daya Guo, Shuo Ren, Junjie Huang, Alexey Svyatkovskiy, Ambrosio
  Blanco, Colin Clement, Dawn Drain, Daxin Jiang, Duyu Tang, Ge~Li, Lidong
  Zhou, Linjun Shou, Long Zhou, Michele Tufano, Ming Gong, Ming Zhou, Nan Duan,
  Neel Sundaresan, Shao~Kun Deng, Shengyu Fu, and Shujie Liu.
\newblock Codexglue: A machine learning benchmark dataset for code
  understanding and generation, 2021.
\newblock URL \url{https://arxiv.org/abs/2102.04664}.

\bibitem[Mitchell et~al.(2011)Mitchell, O'Sullivan, and
  Dunning]{Mitchell2011PuLPA}
Stuart Mitchell, Michael~J. O'Sullivan, and Iain Dunning.
\newblock Pulp : A linear programming toolkit for python.
\newblock In \emph{PuLP : A Linear Programming Toolkit for Python}, 2011.

\bibitem[Nye et~al.(2021)Nye, Andreassen, Gur-Ari, Michalewski, Austin, Bieber,
  Dohan, Lewkowycz, Bosma, Luan, Sutton, and Odena]{scratchpad}
Maxwell Nye, Anders~Johan Andreassen, Guy Gur-Ari, Henryk Michalewski, Jacob
  Austin, David Bieber, David Dohan, Aitor Lewkowycz, Maarten Bosma, David
  Luan, Charles Sutton, and Augustus Odena.
\newblock Show your work: Scratchpads for intermediate computation with
  language models, 2021.
\newblock URL \url{https://arxiv.org/abs/2112.00114}.

\bibitem[Ofoghi et~al.(2020)Ofoghi, Mak, and
  Yearwood]{https://doi.org/10.48550/arxiv.2011.06300}
Bahadorreza Ofoghi, Vicky Mak, and John Yearwood.
\newblock A knowledge representation approach to automated mathematical
  modelling, 2020.
\newblock URL \url{https://arxiv.org/abs/2011.06300}.

\bibitem[Parmentier(2021)]{https://doi.org/10.48550/arxiv.2107.04323}
Axel Parmentier.
\newblock Learning structured approximations of combinatorial optimization
  problems, 2021.
\newblock URL \url{https://arxiv.org/abs/2107.04323}.

\bibitem[Poesia et~al.(2022)Poesia, Polozov, Le, Tiwari, Soares, Meek, and
  Gulwani]{synchromesh}
Gabriel Poesia, Oleksandr Polozov, Vu~Le, Ashish Tiwari, Gustavo Soares,
  Christopher Meek, and Sumit Gulwani.
\newblock Synchromesh: Reliable code generation from pre-trained language
  models, 2022.
\newblock URL \url{https://arxiv.org/abs/2201.11227}.

\bibitem[Ramamonjison et~al.(2022{\natexlab{a}})Ramamonjison, Li, Yu, He,
  Rengan, Banitalebi-Dehkordi, Zhou, and Zhang]{NL4OPT}
Rindranirina Ramamonjison, Haley Li, Timothy~T. Yu, Shiqi He, Vishnu Rengan,
  Amin Banitalebi-Dehkordi, Zirui Zhou, and Yong Zhang.
\newblock Augmenting operations research with auto-formulation of optimization
  models from problem descriptions, 2022{\natexlab{a}}.
\newblock URL \url{https://arxiv.org/abs/2209.15565}.

\bibitem[Ramamonjison et~al.(2022{\natexlab{b}})Ramamonjison, Li, Yu, He,
  Rengan, Banitalebi-Dehkordi, Zhou, and
  Zhang]{https://doi.org/10.48550/arxiv.2209.15565}
Rindranirina Ramamonjison, Haley Li, Timothy~T. Yu, Shiqi He, Vishnu Rengan,
  Amin Banitalebi-Dehkordi, Zirui Zhou, and Yong Zhang.
\newblock Augmenting operations research with auto-formulation of optimization
  models from problem descriptions, 2022{\natexlab{b}}.
\newblock URL \url{https://arxiv.org/abs/2209.15565}.

\bibitem[Rizk et~al.(2022)Rizk, Venkateswaran, Isahagian, and
  Muthusamy]{https://doi.org/10.48550/arxiv.2210.14739}
Yara Rizk, Praveen Venkateswaran, Vatche Isahagian, and Vinod Muthusamy.
\newblock A case for business process-specific foundation models, 2022.
\newblock URL \url{https://arxiv.org/abs/2210.14739}.

\bibitem[Rossol(1986)]{Rossol1986}
L.~Rossol.
\newblock The {KAREL} language for programmable automation.
\newblock In \emph{Initiativen f\"{u}r die Fabrik mit Zukunft}, pp.\  293--312.
  Springer Berlin Heidelberg, 1986.
\newblock \doi{10.1007/978-3-662-12068-2_18}.
\newblock URL \url{https://doi.org/10.1007/978-3-662-12068-2_18}.

\bibitem[Sahu et~al.(2022)Sahu, Rodriguez, Laradji, Atighehchian, Vazquez, and
  Bahdanau]{DataAug}
Gaurav Sahu, Pau Rodriguez, Issam~H. Laradji, Parmida Atighehchian, David
  Vazquez, and Dzmitry Bahdanau.
\newblock Data augmentation for intent classification with off-the-shelf large
  language models, 2022.
\newblock URL \url{https://arxiv.org/abs/2204.01959}.

\bibitem[Scholak et~al.(2021)Scholak, Schucher, and Bahdanau]{picard}
Torsten Scholak, Nathan Schucher, and Dzmitry Bahdanau.
\newblock Picard: Parsing incrementally for constrained auto-regressive
  decoding from language models, 2021.
\newblock URL \url{https://arxiv.org/abs/2109.05093}.

\bibitem[Van~Hentenryck(1999)]{10.5555/299293}
Pascal Van~Hentenryck.
\newblock \emph{The OPL Optimization Programming Language}.
\newblock MIT Press, Cambridge, MA, USA, 1999.
\newblock ISBN 0262720302.

\bibitem[Wang et~al.(2021)Wang, Wang, Joty, and Hoi]{CodeT5}
Yue Wang, Weishi Wang, Shafiq Joty, and Steven C.~H. Hoi.
\newblock Codet5: Identifier-aware unified pre-trained encoder-decoder models
  for code understanding and generation, 2021.
\newblock URL \url{https://arxiv.org/abs/2109.00859}.

\bibitem[Wei et~al.(2022)Wei, Wang, Schuurmans, Bosma, Ichter, Xia, Chi, Le,
  and Zhou]{Chainofthought}
Jason Wei, Xuezhi Wang, Dale Schuurmans, Maarten Bosma, Brian Ichter, Fei Xia,
  Ed~Chi, Quoc Le, and Denny Zhou.
\newblock Chain-of-thought prompting elicits reasoning in large language
  models, 2022.
\newblock URL \url{https://arxiv.org/abs/2201.11903}.

\bibitem[Workshop et~al.(2022)Workshop, {:}, Scao, Fan, Akiki, Pavlick, Ilić,
  Hesslow, Castagné, Luccioni, Yvon, Gallé, Tow, Rush, Biderman, Webson,
  Ammanamanchi, Wang, Sagot, Muennighoff, del Moral, Ruwase, Bawden, Bekman,
  McMillan-Major, Beltagy, Nguyen, Saulnier, Tan, Suarez, Sanh, Laurençon,
  Jernite, Launay, Mitchell, Raffel, Gokaslan, Simhi, Soroa, Aji, Alfassy,
  Rogers, Nitzav, Xu, Mou, Emezue, Klamm, Leong, van Strien, Adelani, Radev,
  Ponferrada, Levkovizh, Kim, Natan, De~Toni, Dupont, Kruszewski, Pistilli,
  Elsahar, Benyamina, Tran, Yu, Abdulmumin, Johnson, Gonzalez-Dios, de~la Rosa,
  Chim, Dodge, Zhu, Chang, Frohberg, Tobing, Bhattacharjee, Almubarak, Chen,
  Lo, Von~Werra, Weber, Phan, allal, Tanguy, Dey, Muñoz, Masoud, Grandury,
  Šaško, Huang, Coavoux, Singh, Jiang, Vu, Jauhar, Ghaleb, Subramani,
  Kassner, Khamis, Nguyen, Espejel, de~Gibert, Villegas, Henderson, Colombo,
  Amuok, Lhoest, Harliman, Bommasani, López, Ribeiro, Osei, Pyysalo, Nagel,
  Bose, Muhammad, Sharma, Longpre, Nikpoor, Silberberg, Pai, Zink, Torrent,
  Schick, Thrush, Danchev, Nikoulina, Laippala, Lepercq, Prabhu, Alyafeai,
  Talat, Raja, Heinzerling, Si, Taşar, Salesky, Mielke, Lee, Sharma, Santilli,
  Chaffin, Stiegler, Datta, Szczechla, Chhablani, Wang, Pandey, Strobelt,
  Fries, Rozen, Gao, Sutawika, Bari, Al-shaibani, Manica, Nayak, Teehan,
  Albanie, Shen, Ben-David, Bach, Kim, Bers, Fevry, Neeraj, Thakker, Raunak,
  Tang, Yong, Sun, Brody, Uri, Tojarieh, Roberts, Chung, Tae, Phang, Press, Li,
  Narayanan, Bourfoune, Casper, Rasley, Ryabinin, Mishra, Zhang, Shoeybi,
  Peyrounette, Patry, Tazi, Sanseviero, von Platen, Cornette, Lavallée,
  Lacroix, Rajbhandari, Gandhi, Smith, Requena, Patil, Dettmers, Baruwa, Singh,
  Cheveleva, Ligozat, Subramonian, Névéol, Lovering, Garrette, Tunuguntla,
  Reiter, Taktasheva, Voloshina, Bogdanov, Winata, Schoelkopf, Kalo, Novikova,
  Forde, Clive, Kasai, Kawamura, Hazan, Carpuat, Clinciu, Kim, Cheng, Serikov,
  Antverg, van~der Wal, Zhang, Zhang, Gehrmann, Mirkin, Pais, Shavrina,
  Scialom, Yun, Limisiewicz, Rieser, Protasov, Mikhailov, Pruksachatkun,
  Belinkov, Bamberger, Kasner, Rueda, Pestana, Feizpour, Khan, Faranak, Santos,
  Hevia, Unldreaj, Aghagol, Abdollahi, Tammour, HajiHosseini, Behroozi,
  Ajibade, Saxena, Ferrandis, Contractor, Lansky, David, Kiela, Nguyen, Tan,
  Baylor, Ozoani, Mirza, Ononiwu, Rezanejad, Jones, Bhattacharya, Solaiman,
  Sedenko, Nejadgholi, Passmore, Seltzer, Sanz, Dutra, Samagaio, Elbadri,
  Mieskes, Gerchick, Akinlolu, McKenna, Qiu, Ghauri, Burynok, Abrar, Rajani,
  Elkott, Fahmy, Samuel, An, Kromann, Hao, Alizadeh, Shubber, Wang, Roy,
  Viguier, Le, Oyebade, Le, Yang, Nguyen, Kashyap, Palasciano, Callahan,
  Shukla, Miranda-Escalada, Singh, Beilharz, Wang, Brito, Zhou, Jain, Xu,
  Fourrier, Periñán, Molano, Yu, Manjavacas, Barth, Fuhrimann, Altay, Bayrak,
  Burns, Vrabec, Bello, Dash, Kang, Giorgi, Golde, Posada, Sivaraman,
  Bulchandani, Liu, Shinzato, de~Bykhovetz, Takeuchi, Pàmies, Castillo,
  Nezhurina, Sänger, Samwald, Cullan, Weinberg, De~Wolf, Mihaljcic, Liu,
  Freidank, Kang, Seelam, Dahlberg, Broad, Muellner, Fung, Haller,
  Chandrasekhar, Eisenberg, Martin, Canalli, Su, Su, Cahyawijaya, Garda,
  Deshmukh, Mishra, Kiblawi, Ott, Sang-aroonsiri, Kumar, Schweter, Bharati,
  Laud, Gigant, Kainuma, Kusa, Labrak, Bajaj, Venkatraman, Xu, Xu, Xu, Tan,
  Xie, Ye, Bras, Belkada, and Wolf]{Bloom}
BigScience Workshop, {:}, Teven~Le Scao, Angela Fan, Christopher Akiki, Ellie
  Pavlick, Suzana Ilić, Daniel Hesslow, Roman Castagné, Alexandra~Sasha
  Luccioni, François Yvon, Matthias Gallé, Jonathan Tow, Alexander~M. Rush,
  Stella Biderman, Albert Webson, Pawan~Sasanka Ammanamanchi, Thomas Wang,
  Benoît Sagot, Niklas Muennighoff, Albert~Villanova del Moral, Olatunji
  Ruwase, Rachel Bawden, Stas Bekman, Angelina McMillan-Major, Iz~Beltagy, Huu
  Nguyen, Lucile Saulnier, Samson Tan, Pedro~Ortiz Suarez, Victor Sanh, Hugo
  Laurençon, Yacine Jernite, Julien Launay, Margaret Mitchell, Colin Raffel,
  Aaron Gokaslan, Adi Simhi, Aitor Soroa, Alham~Fikri Aji, Amit Alfassy, Anna
  Rogers, Ariel~Kreisberg Nitzav, Canwen Xu, Chenghao Mou, Chris Emezue,
  Christopher Klamm, Colin Leong, Daniel van Strien, David~Ifeoluwa Adelani,
  Dragomir Radev, Eduardo~González Ponferrada, Efrat Levkovizh, Ethan Kim,
  Eyal~Bar Natan, Francesco De~Toni, Gérard Dupont, Germán Kruszewski, Giada
  Pistilli, Hady Elsahar, Hamza Benyamina, Hieu Tran, Ian Yu, Idris Abdulmumin,
  Isaac Johnson, Itziar Gonzalez-Dios, Javier de~la Rosa, Jenny Chim, Jesse
  Dodge, Jian Zhu, Jonathan Chang, Jörg Frohberg, Joseph Tobing, Joydeep
  Bhattacharjee, Khalid Almubarak, Kimbo Chen, Kyle Lo, Leandro Von~Werra, Leon
  Weber, Long Phan, Loubna~Ben allal, Ludovic Tanguy, Manan Dey, Manuel~Romero
  Muñoz, Maraim Masoud, María Grandury, Mario Šaško, Max Huang, Maximin
  Coavoux, Mayank Singh, Mike Tian-Jian Jiang, Minh~Chien Vu, Mohammad~A.
  Jauhar, Mustafa Ghaleb, Nishant Subramani, Nora Kassner, Nurulaqilla Khamis,
  Olivier Nguyen, Omar Espejel, Ona de~Gibert, Paulo Villegas, Peter Henderson,
  Pierre Colombo, Priscilla Amuok, Quentin Lhoest, Rheza Harliman, Rishi
  Bommasani, Roberto~Luis López, Rui Ribeiro, Salomey Osei, Sampo Pyysalo,
  Sebastian Nagel, Shamik Bose, Shamsuddeen~Hassan Muhammad, Shanya Sharma,
  Shayne Longpre, Somaieh Nikpoor, Stanislav Silberberg, Suhas Pai, Sydney
  Zink, Tiago~Timponi Torrent, Timo Schick, Tristan Thrush, Valentin Danchev,
  Vassilina Nikoulina, Veronika Laippala, Violette Lepercq, Vrinda Prabhu, Zaid
  Alyafeai, Zeerak Talat, Arun Raja, Benjamin Heinzerling, Chenglei Si,
  Davut~Emre Taşar, Elizabeth Salesky, Sabrina~J. Mielke, Wilson~Y. Lee,
  Abheesht Sharma, Andrea Santilli, Antoine Chaffin, Arnaud Stiegler, Debajyoti
  Datta, Eliza Szczechla, Gunjan Chhablani, Han Wang, Harshit Pandey, Hendrik
  Strobelt, Jason~Alan Fries, Jos Rozen, Leo Gao, Lintang Sutawika, M~Saiful
  Bari, Maged~S. Al-shaibani, Matteo Manica, Nihal Nayak, Ryan Teehan, Samuel
  Albanie, Sheng Shen, Srulik Ben-David, Stephen~H. Bach, Taewoon Kim, Tali
  Bers, Thibault Fevry, Trishala Neeraj, Urmish Thakker, Vikas Raunak, Xiangru
  Tang, Zheng-Xin Yong, Zhiqing Sun, Shaked Brody, Yallow Uri, Hadar Tojarieh,
  Adam Roberts, Hyung~Won Chung, Jaesung Tae, Jason Phang, Ofir Press, Conglong
  Li, Deepak Narayanan, Hatim Bourfoune, Jared Casper, Jeff Rasley, Max
  Ryabinin, Mayank Mishra, Minjia Zhang, Mohammad Shoeybi, Myriam Peyrounette,
  Nicolas Patry, Nouamane Tazi, Omar Sanseviero, Patrick von Platen, Pierre
  Cornette, Pierre~François Lavallée, Rémi Lacroix, Samyam Rajbhandari,
  Sanchit Gandhi, Shaden Smith, Stéphane Requena, Suraj Patil, Tim Dettmers,
  Ahmed Baruwa, Amanpreet Singh, Anastasia Cheveleva, Anne-Laure Ligozat, Arjun
  Subramonian, Aurélie Névéol, Charles Lovering, Dan Garrette, Deepak
  Tunuguntla, Ehud Reiter, Ekaterina Taktasheva, Ekaterina Voloshina, Eli
  Bogdanov, Genta~Indra Winata, Hailey Schoelkopf, Jan-Christoph Kalo,
  Jekaterina Novikova, Jessica~Zosa Forde, Jordan Clive, Jungo Kasai, Ken
  Kawamura, Liam Hazan, Marine Carpuat, Miruna Clinciu, Najoung Kim, Newton
  Cheng, Oleg Serikov, Omer Antverg, Oskar van~der Wal, Rui Zhang, Ruochen
  Zhang, Sebastian Gehrmann, Shachar Mirkin, Shani Pais, Tatiana Shavrina,
  Thomas Scialom, Tian Yun, Tomasz Limisiewicz, Verena Rieser, Vitaly Protasov,
  Vladislav Mikhailov, Yada Pruksachatkun, Yonatan Belinkov, Zachary Bamberger,
  Zdeněk Kasner, Alice Rueda, Amanda Pestana, Amir Feizpour, Ammar Khan, Amy
  Faranak, Ana Santos, Anthony Hevia, Antigona Unldreaj, Arash Aghagol, Arezoo
  Abdollahi, Aycha Tammour, Azadeh HajiHosseini, Bahareh Behroozi, Benjamin
  Ajibade, Bharat Saxena, Carlos~Muñoz Ferrandis, Danish Contractor, David
  Lansky, Davis David, Douwe Kiela, Duong~A. Nguyen, Edward Tan, Emi Baylor,
  Ezinwanne Ozoani, Fatima Mirza, Frankline Ononiwu, Habib Rezanejad, Hessie
  Jones, Indrani Bhattacharya, Irene Solaiman, Irina Sedenko, Isar Nejadgholi,
  Jesse Passmore, Josh Seltzer, Julio~Bonis Sanz, Livia Dutra, Mairon Samagaio,
  Maraim Elbadri, Margot Mieskes, Marissa Gerchick, Martha Akinlolu, Michael
  McKenna, Mike Qiu, Muhammed Ghauri, Mykola Burynok, Nafis Abrar, Nazneen
  Rajani, Nour Elkott, Nour Fahmy, Olanrewaju Samuel, Ran An, Rasmus Kromann,
  Ryan Hao, Samira Alizadeh, Sarmad Shubber, Silas Wang, Sourav Roy, Sylvain
  Viguier, Thanh Le, Tobi Oyebade, Trieu Le, Yoyo Yang, Zach Nguyen,
  Abhinav~Ramesh Kashyap, Alfredo Palasciano, Alison Callahan, Anima Shukla,
  Antonio Miranda-Escalada, Ayush Singh, Benjamin Beilharz, Bo~Wang, Caio
  Brito, Chenxi Zhou, Chirag Jain, Chuxin Xu, Clémentine Fourrier,
  Daniel~León Periñán, Daniel Molano, Dian Yu, Enrique Manjavacas, Fabio
  Barth, Florian Fuhrimann, Gabriel Altay, Giyaseddin Bayrak, Gully Burns,
  Helena~U. Vrabec, Imane Bello, Ishani Dash, Jihyun Kang, John Giorgi, Jonas
  Golde, Jose~David Posada, Karthik~Rangasai Sivaraman, Lokesh Bulchandani,
  Lu~Liu, Luisa Shinzato, Madeleine~Hahn de~Bykhovetz, Maiko Takeuchi, Marc
  Pàmies, Maria~A Castillo, Marianna Nezhurina, Mario Sänger, Matthias
  Samwald, Michael Cullan, Michael Weinberg, Michiel De~Wolf, Mina Mihaljcic,
  Minna Liu, Moritz Freidank, Myungsun Kang, Natasha Seelam, Nathan Dahlberg,
  Nicholas~Michio Broad, Nikolaus Muellner, Pascale Fung, Patrick Haller, Ramya
  Chandrasekhar, Renata Eisenberg, Robert Martin, Rodrigo Canalli, Rosaline Su,
  Ruisi Su, Samuel Cahyawijaya, Samuele Garda, Shlok~S Deshmukh, Shubhanshu
  Mishra, Sid Kiblawi, Simon Ott, Sinee Sang-aroonsiri, Srishti Kumar, Stefan
  Schweter, Sushil Bharati, Tanmay Laud, Théo Gigant, Tomoya Kainuma, Wojciech
  Kusa, Yanis Labrak, Yash~Shailesh Bajaj, Yash Venkatraman, Yifan Xu, Yingxin
  Xu, Yu~Xu, Zhe Tan, Zhongli Xie, Zifan Ye, Mathilde Bras, Younes Belkada, and
  Thomas Wolf.
\newblock Bloom: A 176b-parameter open-access multilingual language model,
  2022.
\newblock URL \url{https://arxiv.org/abs/2211.05100}.

\bibitem[Yoo et~al.(2021)Yoo, Park, Kang, Lee, and Park]{GPT3Mix}
Kang~Min Yoo, Dongju Park, Jaewook Kang, Sang-Woo Lee, and Woomyeong Park.
\newblock Gpt3mix: Leveraging large-scale language models for text
  augmentation, 2021.
\newblock URL \url{https://arxiv.org/abs/2104.08826}.

\bibitem[Zhong et~al.(2017)Zhong, Xiong, and Socher]{seq2sql}
Victor Zhong, Caiming Xiong, and Richard Socher.
\newblock Seq2sql: Generating structured queries from natural language using
  reinforcement learning, 2017.
\newblock URL \url{https://arxiv.org/abs/1709.00103}.

\end{thebibliography}
